\documentclass[sigconf]{acmart}
\AtBeginDocument{%
  }

\setcopyright{acmlicensed}
\copyrightyear{2018}
\acmYear{2018}
\acmDOI{XXXXXXX.XXXXXXX}
\acmConference[CCS ’25]{Make sure to enter the correct
  conference title from your rights confirmation email}
\acmISBN{978-1-4503-XXXX-X/2018/06}

\acmSubmissionID{1787}

\usepackage{multirow}
\usepackage[normalem]{ulem}
\useunder{\uline}{\ul}{}
\usepackage{comment}
\usepackage{booktabs}

\usepackage{algorithm}
\usepackage{algpseudocode}
\usepackage{amsmath}
\usepackage{graphicx}
\usepackage{array}

\begin{document}

\title[RoVo: Robust Voice Protection]{RoVo: Robust Voice Protection Against Unauthorized Speech Synthesis with Embedding-Level Perturbations}


\author{Seungmin Kim}
\email{kimsm0802@soongsil.ac.kr}
\orcid{0009-0002-0911-3282}
\authornotemark[1]
\affiliation{%
  \institution{Soongsil University}
  \city{seoul}
  \country{South Korea}
}

\author{Sohee Park}
\email{sosohi@soongsil.ac.kr}
\orcid{0009-0005-4897-7579}
\affiliation{%
  \institution{Soongsil University}
  \city{seoul}
  \country{South Korea}
}

\author{Donghyun Kim}
\email{kmit16@soongsil.ac.kr}
\orcid{0009-0006-2033-4341}
\affiliation{%
  \institution{Soongsil University}
  \city{seoul}
  \country{South Korea}
}

\author{Jisu Lee}
\email{connandgo@soongsil.ac.kr}
\orcid{0009-0009-8932-6333}
\affiliation{%
  \institution{Soongsil University}
  \city{seoul}
  \country{South Korea}
}

\author{Daeseon Choi}
\email{sunchoi@ssu.ac.kr}
\orcid{0000-0002-1438-0265}
\authornotemark[1]
\affiliation{%
  \institution{Soongsil University}
  \city{seoul}
  \country{South Korea}
}

\renewcommand{\shortauthors}{s.Kim et al.}

\begin{abstract}
With the advancement of AI-based speech synthesis technologies such as Deep Voice, there is an increasing risk of voice spoofing attacks, including voice phishing and fake news, through unauthorized use of others' voices. Existing defenses that inject adversarial perturbations directly into audio signals have limited effectiveness, as these perturbations can easily be neutralized by speech enhancement methods. To overcome this limitation, we propose \textbf{RoVo (Robust Voice)}\footnote{Sound samples available at: \url{https://smerge0802.github.io/RoVo/}}, a novel proactive defense technique that injects adversarial perturbations into high-dimensional embedding vectors of audio signals, reconstructing them into protected speech. This approach effectively defends against speech synthesis attacks and also provides strong resistance to speech enhancement models, which represent a secondary attack threat.

In extensive experiments, RoVo increased the Defense Success Rate (DSR) by over 70\% compared to unprotected speech, across four state-of-the-art speech synthesis models. Specifically, RoVo achieved a DSR of 99.5\% on a commercial speaker-verification API, effectively neutralizing speech synthesis attack. Moreover, RoVo's perturbations remained robust even under strong speech enhancement conditions, outperforming traditional methods. A user study confirmed that RoVo preserves both naturalness and usability of protected speech, highlighting its effectiveness in complex and evolving threat scenarios.
\end{abstract}

\ccsdesc[500]{Computing methodologies~Machine learning}
\ccsdesc[500]{Security and privacy~Human and societal aspects of security and privacy}

\keywords{Embedding-Level Adversarial Perturbation, DeepVoice Defense, Speech Synthesis}


\maketitle

\section{Introduction}

\begin{figure*}[t]
\centering
\includegraphics[width=\textwidth]{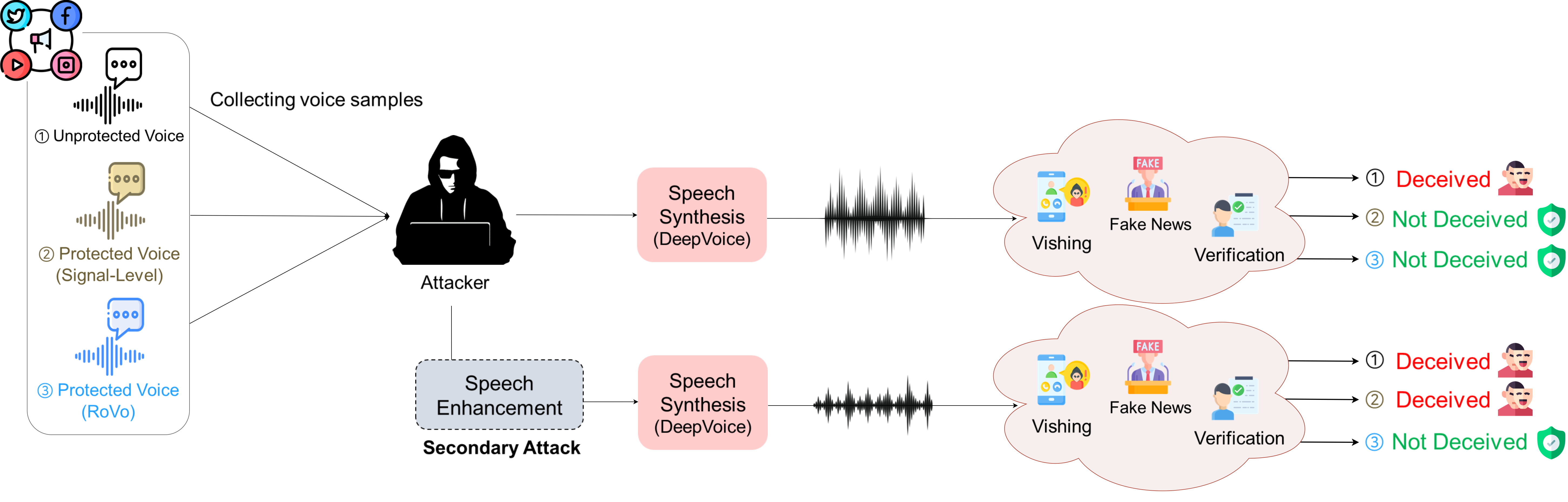}
\caption{\textbf{Overview of RoVo: While signal-level protected audio can defend against the initial attack, its effectiveness significantly decreases under realistic secondary post-processing attacks (e.g., \emph{speech enhancement}). In contrast, RoVo maintains robust defense even against such post-processing, ultimately influencing whether the victim is Deceived or Not Deceived.}} 
\label{fig_1}
\end{figure*}

With the rapid advancements in large-scale datasets and AI (artificial intelligence) algorithms, AI-based data generation technologies have significantly progressed in recent years. Particularly, the fields of speech and image synthesis have achieved remarkable improvements, enabling the creation of synthetic data nearly indistinguishable from real data. These technological advancements present numerous application opportunities across various industries and domains. Among these, TTS (Text-to-Speech) technology, which converts text into natural and personalized speech, has attracted considerable attention due to its wide-ranging uses in personalized voice services, conversational AI, and medical assistive tools. Moreover, recent developments in deep learning technologies, such as Deep Voice, have made it possible to generate highly realistic and emotionally expressive voices even from minimal data. Additionally, VC (Voice Conversion) techniques have substantially advanced, enabling the transformation of one speaker’s voice into that of another, effectively capturing the target speaker’s vocal characteristics while maintaining naturalness. These advancements are expanding the scope of voice data applications and laying the foundation for an era of personalized voice content.\\

\noindent\textbf{The Malicious Uses of AI Speech Synthesis.} While AI-based speech synthesis technology brings numerous positive advancements, it also introduces significant security risks due to its potential misuse. In particular, the capability to accurately replicate an individual's voice to generate speech that the person never actually uttered poses severe threats, such as generating fake news or conducting voice phishing attacks.

For example, in October 2024, criminals in California, USA, used AI-based speech synthesis to precisely replicate a victim’s son's voice, falsely claiming he had been involved in a car accident and urgently needed bail money. Consequently, they successfully deceived the victim, stealing \$25,000 \cite{NYPost2024}. Similarly,in February 2025, Italian authorities froze funds linked to an AI-powered voice scam targeting business executives, where attackers impersonated an official to demand urgent financial support \cite{Reuters2025}. These incidents illustrate severe consequences that go beyond individual harm, raising concerns about political manipulation, societal instability, and the rapid dissemination of misinformation \cite{zhadan2023, cox2023ai}.\\

\noindent \textbf{Limitations and Challenges of Proactive Defense.} With malicious uses of AI-based speech synthesis increasing, effective proactive defense methods have become essential. Current defensive approaches, such as deep neural network-based synthetic-speech detectors \cite{blue2022you, wang2020deepsonar} and audio watermarking \cite{juvela2024collaborative, singh2024silentcipher}, mostly focus on verifying authenticity after content generation. These techniques thus inherently lack the ability to proactively prevent malicious speech synthesis. Given the ease of collecting voice samples from platforms like social media or YouTube, relying solely on post-generation detection is insufficient. Therefore, developing proactive measures that fundamentally protect voice data and prevent malicious synthesis from occurring is crucial.

A promising method to proactively prevent unauthorized speech synthesis involves adversarial perturbation techniques. These methods inject subtle and carefully designed perturbations into audio signals to disrupt the extraction of key voice features. Consequently, synthesized audio becomes distorted or resembles a different speaker, making it difficult for speech synthesis models to accurately reproduce the original speaker’s voice. However, current adversarial perturbation approaches mainly add perturbations directly at the audio signal level. This approach suffers from a fundamental weakness: perturbations can be easily neutralized by widely used speech enhancement \cite{das2021fundamentals} or filtering techniques. In fact, experiments with state-of-the-art adversarial defense methods, such as Antifake \cite{yu2023antifake}, have demonstrated that speech enhancement techniques effectively remove injected perturbations, substantially reducing defense effectiveness. Therefore, it is essential to develop novel defensive methods capable of overcoming these critical vulnerabilities to reliably protect voice data against synthesis attacks. \\

\noindent \textbf{RoVo.} we propose RoVo (Robust Voice), a novel proactive defense mechanism designed to prevent unauthorized speech synthesis attacks while maintaining strong robustness against secondary attacks such as speech enhancement. Existing adversarial perturbation methods typically insert perturbations directly at the signal level. However, such perturbations are easily removed by standard speech enhancement techniques, significantly reducing their defensive effectiveness. To overcome this limitation, RoVo injects adversarial perturbations into high-dimensional embedding representations of audio signals, rather than directly into the signals themselves. By embedding perturbations into the internal representation space, RoVo effectively alters complex voice features, providing stronger resilience against  speech enhancement methods compared to traditional signal-level defenses.

Our experimental results demonstrate that RoVo achieves high Defense Success Rates (DSR) against various speech synthesis models, including challenging black-box scenarios. Notably, under speech enhancement conditions, RoVo substantially outperforms Antifake, a state-of-the-art adversarial defense, confirming its superior robustness. Furthermore, a user study confirmed that RoVo preserves naturalness and usability of protected speech, highlighting its practical applicability. Overall, RoVo represents a practical and effective defensive approach capable of robustly protecting voice data from unauthorized speech synthesis attacks and subsequent enhancement-based secondary attacks.

\noindent \textbf{The contributions of this paper are as follows:}
\begin{itemize}
    \item \textbf{Proactive Defense:} We propose RoVo (Robust Voice), a proactive defense method injecting adversarial perturbations at the embedding level to effectively prevent unauthorized speech synthesis attacks.
    
    \item \textbf{Robustness Against Speech Enhancement Attacks:} RoVo exhibits strong robustness against speech enhancement attacks, significantly outperforming signal-level adversarial perturbation under realistic attack scenarios.

    \item \textbf{Effectiveness and Practicality:} RoVo achieves superior defense effectiveness, improving DSR by over 70\% compared to unprotected speech, and attaining a 99.5\% DSR when measured using a commercial speaker-verification API. User studies further confirm RoVo maintains naturalness and usability, highlighting its practicality.

\end{itemize}

\noindent The remainder of this paper is structured as follows. Section \ref{sec 2} introduces speech synthesis attacks and existing defenses, analyzing their limitations. Section \ref{sec 3} defines the threat model, specifying attacker capabilities and objectives, and establishes RoVo’s defense goals. Section \ref{sec 4} explains the RoVo defense technique using adversarial perturbations at the voice embedding level. Section \ref{sec 5} describes the experimental setup for performance evaluation, and Section \ref{sec 6} discusses RoVo’s defensive performance and robustness based on experimental results. Section \ref{sec 7} summarizes key findings and limitations, and Section \ref{sec 8} concludes with directions for future research.

\section{Background and Related work}\label{sec 2}
\subsection{Speech Synthesis}
Recent advancements in deep learning have significantly improved speech synthesis technologies, enabling the generation of highly realistic synthetic speech that is nearly indistinguishable from authentic human voices. Two representative technologies, VC (Voice Conversion) \cite{huang2021far} and TTS (Text-to-Speech) \cite{dutoit1997introduction}, have been widely adopted in diverse applications such as personalized voice services, audiobooks, dubbing, and conversational AI. However, these technologies also pose serious security threats, including fake news dissemination, sophisticated voice phishing attacks, public opinion manipulation, and bypassing speaker verification systems \cite{cai2023identifying, abdullah2021hear}. 

\begin{figure}[t]
\centering
\includegraphics[width=\linewidth]{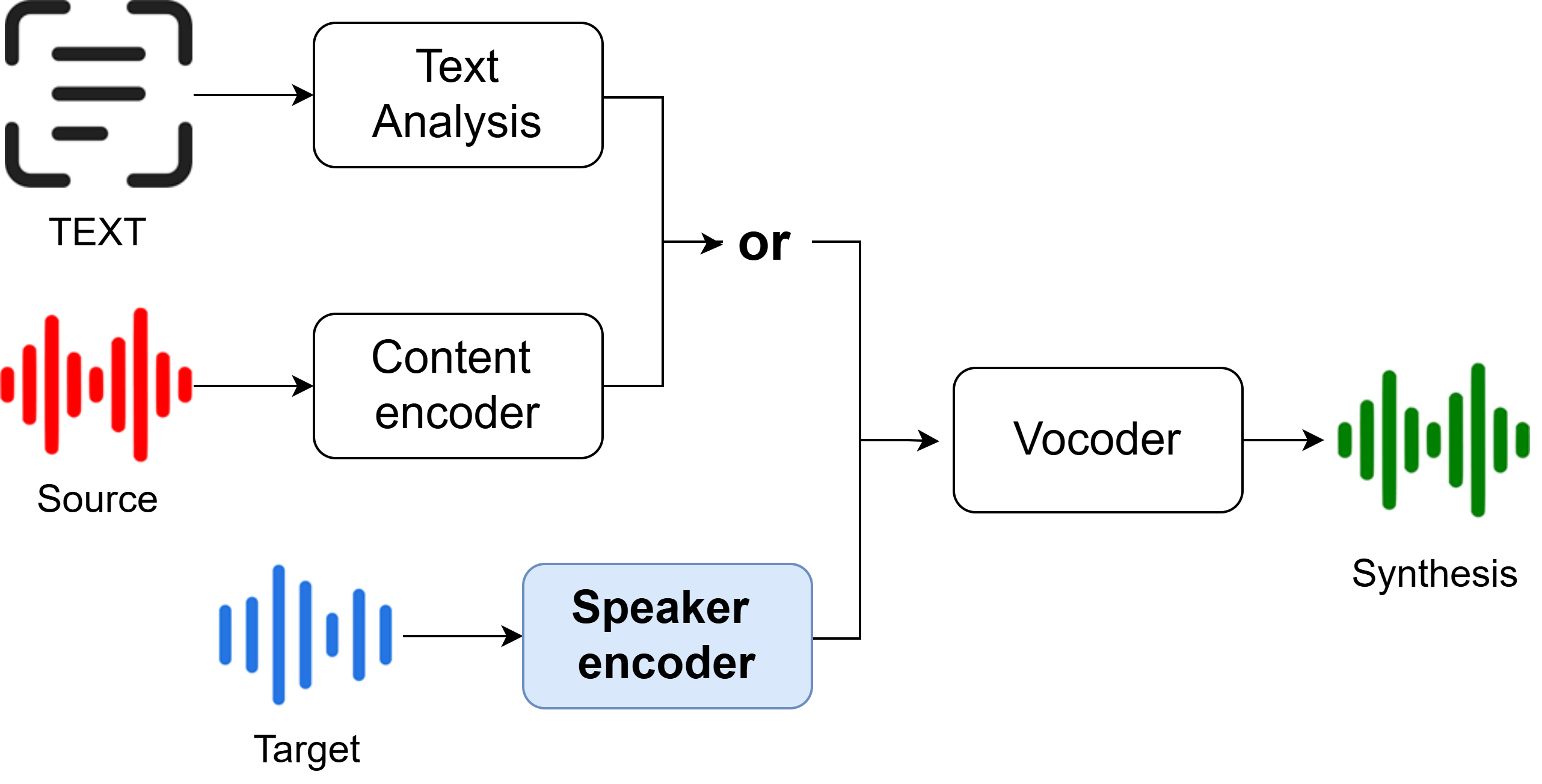}
\caption{\textbf{Architecture of speech synthesis}}
\label{fig_2}
\end{figure}

The architecture of speech synthesis is illustrated in Figure \ref{fig_2}. \\

\noindent\textbf{VC (Voice Conversion)} \cite{qian2019autovc, sisman2020overview} transforms the voice of a source speaker into one that closely matches the target speaker’s vocal attributes, such as pitch, timbre, and prosody, while preserving the original linguistic content. Typically, VC models utilize a \textit{content encoder} to extract linguistic features including phonetic structure and rhythm from the source audio. Concurrently, a \textit{speaker encoder} captures distinctive vocal characteristics from the target speaker’s audio samples. These two feature sets are subsequently integrated through a neural-based \textit{vocoder}, synthesizing speech waveforms that precisely replicate the target speaker’s voice. Consequently, VC models can convincingly replicate a speaker’s voice identity, even from limited audio samples.\\

\noindent\textbf{TTS (Text-to-Speech)} \cite{ren2019fastspeech, casanova2022yourtts, kong2020hifi, jia2018transfer} generates natural, intelligible, and highly realistic speech directly from textual input. Typically, TTS models begin by analyzing input text through a \textit{text analysis} module, extracting critical linguistic details such as phonemes, duration, intonation, and stress patterns. In parallel, a \textit{speaker encoder} obtains voice-specific embeddings by analyzing audio samples from the target speaker. Finally, the linguistic and speaker-specific embeddings are fused within a high-quality neural \textit{vocoder} to synthesize speech closely resembling the target speaker’s vocal attributes and speaking style. Recent deep-learning-based TTS architectures have demonstrated remarkable capabilities in generating natural, expressive, and human-like speech. \\

\noindent\textbf{Speaker Encoder.} A crucial component influencing the performance of speech synthesis systems—especially those designed to replicate specific speaker identities—is the \textit{speaker embedding} \cite{chou2019one, wan2018generalized, heo2020clova, alexander2021image}, typically represented as a fixed-length vector capturing the unique vocal identity of individual speakers. An effective embedding space maintains consistently high similarity among embeddings extracted from the same speaker, while clearly separating embeddings obtained from different speakers. Recent speech synthesis techniques typically segment audio signals and mel-spectrograms into smaller units, thereby allowing more precise capture of phonetic and prosodic nuances and significantly reducing inaccuracies arising from averaging over long temporal durations.

Motivated by the importance of the speaker encoder in synthesizing target voices, this study proactively disrupts its ability to extract speaker characteristics, thereby mitigating threats from unauthorized speech synthesis attacks.

\subsection{Defense Against DeepVoice}
With the rapid development of speech synthesis technologies such as DeepVoice, research on defending against related attacks has gained significant momentum. Most existing approaches primarily focus on detecting \cite{yi2023audio} synthetic speech by analyzing speaker characteristics or acoustic cues to assess its “liveness” \cite{wan2018generalized, ahmed2020void, shah2024deepverify, albadawy2019detecting}. However, these methods are highly sensitive to factors like generation conditions and recording devices, limiting their reliability in real-world settings. Although various DNN-based detection models \cite{huang2021far, mutica2024synthetic, yan2024voicewukong} have been proposed and evaluated in challenges like ASVspoof \cite{yamagishi2019asvspoof, delgado2021asvspoof}, their effectiveness and applicability under real-time conditions or genuine adversarial scenarios remain unclear. Furthermore, detection-based approaches inherently cannot prevent the generation of fake speech proactively.

Recent research has shifted towards exploring proactive defense mechanisms to disrupt the speech synthesis process \cite{yu2023antifake, huang2021defending, liu2023protecting}. The primary method proposed thus far involves injecting perturbations into the audio signal to prevent the synthesis model from accurately reflecting the speaker's characteristics. These perturbations cause the synthesized voice to sound different from the original speaker, making it difficult for attackers to replicate the target voice. The Antifake \cite{yu2023antifake} mechanism is a prominent example of this approach. However, a significant drawback of this method is that perturbations can easily be neutralized by speech enhancement technologies. When perturbations are nullified, the defense performance is severely compromised, making it challenging to maintain reliable long-term protection. Additionally, determining the optimal perturbation size poses practical challenges; excessively large perturbations distort the naturalness of the original voice, while overly small perturbations significantly reduce defensive effectiveness. Consequently, current proactive defense techniques face critical practical limitations, underscoring the urgent need for developing more robust and efficient defensive solutions.

\subsection{Speech Enhancement}
Speech enhancement \cite{o2024speech, yuliani2021speech} refers to technologies designed to improve audio quality by removing background noise and distortions, particularly in noisy environments. Its primary objective is to preserve critical speech information, thus enhancing clarity and recognition accuracy. Common applications include telecommunications, conferencing systems, speech recognition, and hearing aids. Key techniques include noise suppression \cite{xu2014regression} to remove environmental noise, voice boosting to enhance speech clarity, and echo cancellation to eliminate reverberation. Recent advancements integrate frequency-domain analysis with deep learning methods \cite{schroter2022deepfilternet, lu2023mp}, adaptively enhancing audio quality without distorting the original content, thereby emphasizing the importance of speech enhancement for preserving clarity and naturalness. However, despite their benefits, these enhancement technologies can inadvertently function as secondary attacks against perturbation-based speech synthesis defense mechanisms. Specifically, the perturbations injected to protect the voice may be neutralized by speech enhancement technologies, compromising the defense's effectiveness. This vulnerability highlights the need for robust defense mechanisms that can withstand the influence of such enhancement techniques.

\noindent \textbf{In this paper}, we address the limitations of existing methods by proposing a robust defense technique that remains resilient against speech enhancement-based secondary attacks, thus providing a practical and reliable solution against speech synthesis threats.

\section{Threat Model}\label{sec 3}
In this section, we define the threat model by specifying the attacker's capabilities and goals, and clearly outline the intended environment and objectives of RoVo users.

\subsection{Attacker's Capabilities and Objectives}

We assume an attacker who possesses advanced AI-driven speech synthesis capabilities, enabling precise imitation or manipulation of specific target speakers' voices. Using state-of-the-art deep learning techniques such as VC and TTS, the attacker can generate highly realistic synthetic speech, virtually indistinguishable from genuine human voices. Leveraging these capabilities, the attacker can pose multiple security threats by unauthorized reproduction of the victim's voice:

\begin{itemize}
    \item \textbf{Voice Phishing:} Attackers replicate the voice of victims or individuals related to the victim to execute sophisticated social-engineering attacks, including financial scams and identity theft.

    \item \textbf{Fake News and Public Opinion Manipulation:} Attackers maliciously synthesize the voices of influential political or public figures to disseminate false information rapidly, aiming to create social confusion or negatively influence public opinion and policy decisions.

    \item \textbf{Bypassing Voice Authentication}: Attackers exploit synthesized voices to bypass voice-authentication systems used in applications such as smartphones, IoT devices, banking services, and secure facility access, enabling unauthorized entry or fraudulent activities.

\end{itemize}

Additionally, we assume attackers will actively employ speech enhancement to neutralize or weaken existing proactive defenses, including RoVo. Specifically, an attacker who obtains protected audio samples might apply advanced speech enhancement methods to attempt removing embedded defensive perturbations.

\subsection{RoVo User}
RoVo users are individuals or organizations adopting this proposed defense mechanism to proactively protect their voice data from unauthorized and malicious speech synthesis. They seek to prevent malicious cloning or misuse of their voices rather than relying on detecting synthesized speech afterward. Specifically, RoVo users adopt this defensive technique under the following objectives and operational assumptions:

\begin{itemize}
    \item \textbf{Proactive Defense:} Users prefer a proactive approach that preemptively blocks unauthorized speech synthesis rather than relying on reactive post-synthesis detection. This addresses inherent limitations of detection-based methods.

    \item \textbf{Robustness in Real-world Environments:} RoVo users require consistent and robust protection across diverse realistic environments, especially in scenarios involving advanced post-processing attacks, including speech enhancement and filtering techniques.

    \item \textbf{Preserving Naturalness and Usability:} Users expect minimal distortion or loss of audio quality, ensuring that the protected speech maintains sufficient clarity, naturalness, and usability in practical, real-world applications.

\end{itemize}
Thus, within our threat model, RoVo users specifically aim for proactive protection against unauthorized speech synthesis, robustness against secondary attacks like speech enhancement, and the preservation of audio naturalness and usability in realistic scenarios.

\begin{figure*}[t]
\centering
\includegraphics[width=0.9\textwidth]{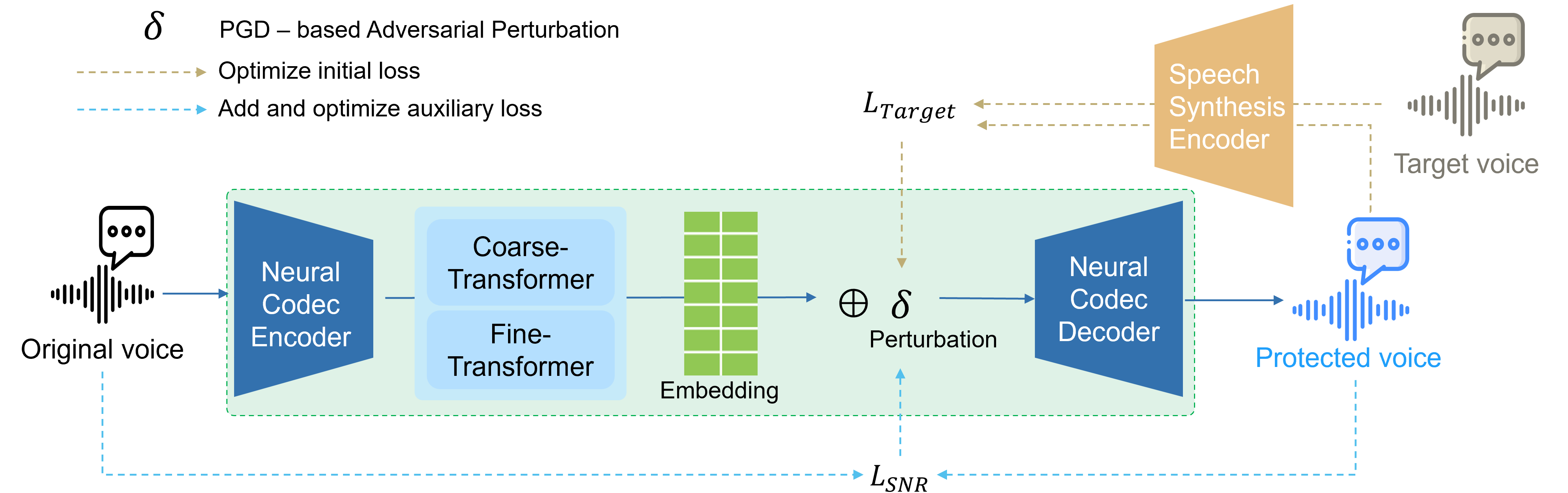}
\caption{Overview of RoVo framework: RoVo injects adversarial perturbations directly into embedding vectors extracted by the Neural Codec Encoder, effectively disrupting speaker-specific features and reconstructing protected audio through the Neural Codec Decoder.}
\label{fig_3}
\end{figure*}

\section{Proposed Method}\label{sec 4}

\subsection{Motivation}
Existing defense methods against synthetic speech attacks primarily rely on injecting adversarial noise directly into audio signals. A representative example is AntiFake
\cite{yu2023antifake}, which adds adversarial perturbations at the signal level to disrupt speaker encoders in speech synthesis models. AntiFake employs two strategies: a target-based approach, manipulating embeddings to mislead the speaker encoder into identifying the speaker incorrectly, and a threshold-based approach, adjusting noise levels to balance defense effectiveness with audio quality. However, our experiments revealed a significant limitation of signal-level methods such as AntiFake. Specifically, we observed that advanced DNN-based speech enhancement models effectively detect and remove signal-level adversarial perturbations, thus neutralizing the defense. After speech enhancement, the synthesized speech retains the characteristics of the original speaker, significantly reducing AntiFake's protective capabilities. This occurs because signal-level perturbations are relatively easy for enhancement algorithms to separate and eliminate.

To address this issue, we propose a novel defense approach that injects adversarial perturbations at the embedding level rather than the signal level. Unlike signal-level noise, embedding-level perturbations persist through speech enhancement processes and remain robust against removal. Consequently, our proposed method provides significantly stronger and more reliable protection against unauthorized speech synthesis attacks, directly addressing the limitations inherent in existing signal-level defenses.

\subsection{Embedding-Level Adversarial Defense}

In this work, we propose RoVo (Robust Voice), a novel voice-defense mechanism that directly injects adversarial perturbations into embedding representations, overcoming limitations of existing signal-level defense approaches.\\

\noindent\textbf{RoVo Overview.} 

\noindent Figure \ref{fig_3} illustrates the overall framework of the RoVo defense mechanism for reconstructing protected speech audio. (1) The original voice input is transformed into a high-dimensional embedding vector via a NAC (Neural Audio Codec) \cite{zeghidour2021soundstream, defossez2022high}. This embedding compactly captures critical speech characteristics, including speaker identity, vocal timbre, and acoustic features. The embedding vector subsequently passes through two transformer-based components: the Coarse Transformer and the Fine Transformer. The Coarse Transformer captures broad, long-term temporal patterns and structural characteristics of speech, while the Fine Transformer refines these embeddings, preserving detailed acoustic features necessary for high-quality, natural-sounding speech reconstruction.

(2) Adversarial perturbations are strategically injected into the transformed embeddings using a PGD (Projected Gradient Descent)-based algorithm \cite{madry2017towards}. By embedding these perturbations directly within high-dimensional speech representations, RoVo disrupts the speaker encoder's ability to accurately extract the original speaker’s unique vocal features. Consequently, unauthorized speech synthesis attempts fail to replicate the original speaker's identity effectively.Crucially, embedding-level perturbations are inherently more difficult to detect and remove than traditional signal-level perturbations, thereby ensuring robust defense even against advanced secondary attacks, such as speech enhancement. (3) Finally, the perturbed embeddings are passed through a NAC Decoder, reconstructing the final protected audio. This step balances security effectiveness and audio quality, maintaining speech clarity and naturalness while resisting unauthorized synthesis attempts.

Overall, RoVo effectively addresses the fundamental limitations of existing signal-level defenses by leveraging embedding-level adversarial perturbations, thus providing robust and stable protection even against sophisticated threats such as advanced speech enhancement-based attacks. \\

\noindent \textbf{Backbone Model.}

\noindent The core concept of RoVo involves transforming speech signals into high-dimensional embedding, strategically injecting adversarial perturbations into these embeddings, and subsequently reconstructing protected audio. For embedding representation and reconstruction effectively, we utilize BARK \cite{barkTTS, panariello2024speaker}\footnote{
Our model was developed using the following open-source repositories: 
Bark TTS (\url{https://github.com/suno-ai/bark}) 
and \texttt{spk\_anon\_nac\_lm} (\url{https://github.com/eurecom-asp/spk_anon_nac_lm}).
}, NAC-based speech synthesis model, as our backbone. BARK is advantageous due to its capability of generating natural and high-quality audio with relatively low computational overhead. Before evaluating RoVo’s defensive performance, we first assessed whether BARK could reliably reconstruct original audio without significant quality degradation.In tests conducted with original audio samples reconstructed via BARK (without perturbations), we achieved a Mean Opinion Score (MOS) of 4.53 and an audio similarity score of 0.96 relative to the original samples. These results confirm that BARK reliably preserves original audio quality and speaker characteristics, making it highly suitable as the backbone model for RoVo.  \\

\begin{figure*}[t]
\centering
\includegraphics[width=\textwidth]{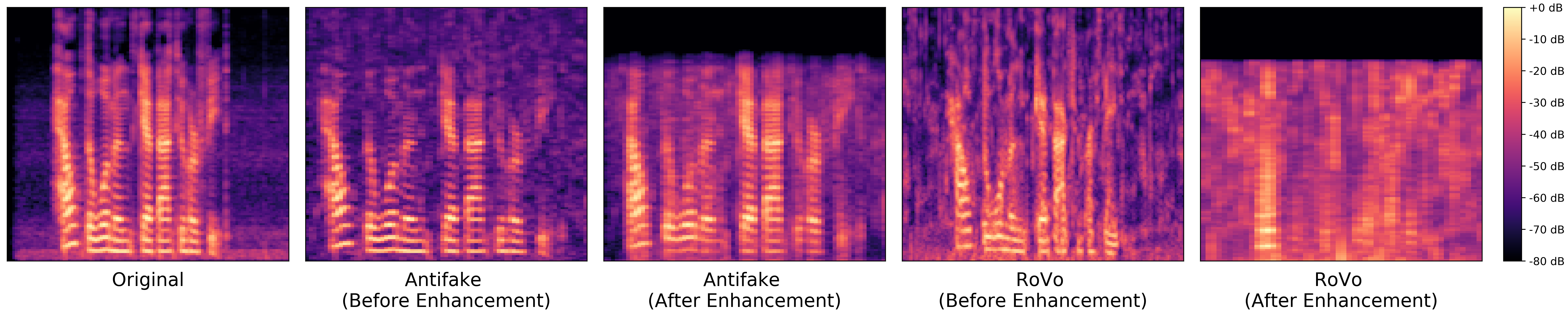}
\caption{Spectrogram comparison: showing Antifake and RoVo before and after speech enhancement. RoVo perturbations remain robust, while Antifake perturbations are easily removed.}
\label{fig_4}
\end{figure*}

\noindent \textbf{PGD-based Adversarial Perturbation.} 

\noindent In this study, we employ a PGD (Projected Gradient Descent)-based algorithm \cite{madry2017towards} to generate such perturbations. Common speech synthesis attack models (e.g., TTS, VC) typically extract the speaker’s unique vocal features through a speaker encoder to generate synthetic speech.Therefore, we specifically design our adversarial perturbations to disrupt accurate extraction of speaker features by the target model’s speaker encoder (detailed in Section \ref{sec 4.3}). To achieve this, we jointly optimize an adversarial loss based on the speaker encoder of the target speech synthesis model, as well as an audio quality preservation loss to maintain the naturalness and perceptual quality of the original speech. This combined approach allows RoVo to generate high-quality audio signals that are simultaneously robust against unauthorized speech synthesis attacks.\\

\noindent\textbf{Robustness Perturbations Against Speech Enhancement.}

\noindent Unlike traditional signal-level defenses, RoVo injects adversarial perturbations directly into high-dimensional embedding representations of audio signals. Existing methods, such as Antifake, apply perturbations directly at the signal level, making these perturbations easily detectable and removable by standard speech enhancement techniques. As demonstrated in Figure \ref{fig_4}, Antifake-generated audio samples exhibit clear signal distortions before enhancement. However, after applying speech enhancement, these distortions are significantly reduced, resulting in audio that closely resembles the original, effectively diminishing the defensive capability.

In contrast, embedding-level perturbations introduced by RoVo generate non-stationary distortions fundamentally different from typical noise patterns learned by enhancement models. Consequently, enhancement models struggle to reliably distinguish RoVo perturbations from genuine speech signals. In our experiments, attempts by speech enhancement methods to remove RoVo's embedding-level perturbations led to significant damage of core speech characteristics, drastically reducing overall audio quality. Specifically, speech signals processed after RoVo protection exhibit abnormal distortions across mid-to-low frequency bands. Spectrogram comparisons further reveal that RoVo-protected audio (before enhancement), despite embedding-level perturbations, clearly retains the original speech’s critical temporal structures and frequency distributions. Notably, RoVo preserves essential phonetic segments and speech features more distinctly compared to Antifake (before enhancement), demonstrating superior audio quality retention.

In summary, RoVo’s embedding-level perturbation strategy introduces novel distortions that conventional speech enhancement models find challenging to mitigate. Thus, RoVo provides robust protection even under sophisticated secondary attacks involving speech enhancement, highlighting the necessity of developing new approaches for detecting and mitigating embedding-level.\\

\subsection{Perturbation Optimization Objectives}\label{sec 4.3}
In this work, we propose to leverage the \textit{Perceptual Alternating Loss (PerC-AL)} framework \cite{zhao2020towards} to simultaneously preserve speech perceptual quality and enhance robustness against unauthorized speech synthesis attacks. PerC-AL dynamically alternates between two complementary loss functions: the \textbf{Target-based Loss}, designed to disrupt speaker embedding extraction, and the \textbf{SNR Loss}, aimed at maintaining audio quality. By adaptively switching these loss objectives according to predefined thresholds, PerC-AL achieves a balanced optimization, ensuring robust protection without perceptible audio degradation.\\

\noindent \textbf{Target-based Loss.} We adopt the Target-based Loss introduced by Antifake~\cite{yu2023antifake}, which distorts speaker embeddings to prevent defended audio from being identified as the original speaker. Specifically, this loss directly operates within the speaker embedding space. Given original and predefined target speaker audio, embeddings are extracted using the speaker encoder of the targeted speech synthesis model. The loss then minimizes the distance between these embeddings. Using this loss within a PGD optimization framework, we iteratively update the adversarial perturbation to be injected into the original speaker's audio. Ultimately, the perturbation disrupts the speaker encoder’s ability to accurately capture the original speaker’s identity, causing speaker verification systems to misclassify the protected audio as a different speaker. Moreover, since this loss directly operates within the embedding space, resulting perturbations inherently resist post-processing attacks, such as speech enhancement, ensuring robust and persistent defense.

The Target-based Loss is formulated as follows:

\begin{equation}
    L_{\text{identity}}(\delta) = D(g(x_U + \delta_{x_U}), g(x_T)),
\end{equation}

where:
\begin{itemize}
    \item \(g(\cdot)\): The speaker embedding extraction encoder.
    \item \(x_U\): The original speaker's audio signal (to be protected).
    \item \(\delta_{x_U}\): The adversarial perturbation specifically crafted.
    \item \(x_T\): The audio signal of a predefined target speaker.
    \item \(D(\cdot, \cdot)\): The distance function measuring the difference.
\end{itemize}

Specifically, \(\delta_{x_U}\) is optimized to minimize the embedding distance between the perturbed original speaker audio \(x_U + \delta_{x_U}\) and the target speaker audio \(x_T\). Consequently, the perturbed audio is misclassified as the target speaker, effectively disrupting unauthorized synthesis attempts.\\

\noindent \textbf{SNR Loss.} The SNR Loss serves as an auxiliary loss to mitigate potential audio quality degradation when relying solely on Target-based Loss for optimization. As the Target-based Loss primarily targets distortion of speaker embeddings, excessively aggressive perturbations may considerably degrade speech quality. To prevent this and effectively limit perturbation magnitudes, we incorporate the SNR Loss. Specifically, this loss maximizes the SNR (signal-to-noise ratio) between the original signal and perturbation, ensuring perturbations remain at appropriate levels. By ensuring the perturbed signal retains characteristics of the original, the SNR Loss crucially balances perceptual quality and robust adversarial defense.\\
\\

The SNR Loss is formulated as follows:

\begin{equation}
    L_{\text{SNR}}(\delta_{x_U}) = 10 \cdot \log_{10} \left(\frac{\|x_U\|^2}{\|\delta_{x_U}\|^2 + \epsilon}\right),
\end{equation}

where:
\begin{itemize}
    \item \(x_U\): The original, unperturbed speech signal to be protected.
    \item \(\delta_{x_U}\): The adversarial perturbation specifically crafted for \(x_U\).
    \item \(\epsilon\): A small constant (\(1 \times 10^{-8}\)) added for numerical stability.
\end{itemize}

Specifically, the SNR Loss maximizes the signal-to-noise ratio between the original speech \(x_U\) and the perturbation \(\delta_{x_U}\), thus effectively constraining perturbation magnitudes and preserving audio perceptual quality.\\

\noindent \textbf{PerC-AL Framework.}
In this study, we propose the PerC-AL (Perceptual Alternating Loss) framework to effectively balance adversarial robustness and perceptual audio quality. PerC-AL dynamically switches between two complementary losses: the Target-based Loss ($L_{\text{identity}}$), which distorts speaker embeddings, and the Signal-to-Noise Ratio (SNR) Loss ($L_{\text{SNR}}$), which preserves audio quality.
Adversarial perturbation-based defenses typically face difficulties simultaneously maintaining audio quality and distorting speaker embeddings effectively. Instead of simultaneously or alternately optimizing both losses, PerC-AL dynamically switches from optimizing the Target-based Loss to the SNR Loss once a predefined threshold is met. Initially, optimization focuses exclusively on the Target-based Loss ($L_{\text{identity}}$) to sufficiently distort speaker embeddings. Once $L_{\text{identity}}$ reaches the predefined threshold ($\tau_{\text{identity}}$), optimization immediately shifts to the SNR Loss ($L_{\text{SNR}}$) to limit perturbation magnitudes and preserve perceptual quality. Optimization terminates when $L_{\text{SNR}}$ stabilizes below its threshold ($\tau_{\text{SNR}}$), yielding the final defended audio. This dynamic loss-switching mechanism ensures efficient and stable training.

The PerC-AL framework is formulated as follows:

\begin{equation}
    L_{\text{PerC-AL}} = 
    \begin{cases} 
        L_{\text{identity}}, & \text{if } L_{\text{identity}} > \tau_{\text{identity}}, \\[6pt]
        L_{\text{SNR}}, & \text{if } L_{\text{identity}} \leq \tau_{\text{identity}} \text{ and } L_{\text{SNR}} > \tau_{\text{SNR}},\\[6pt]
        0, & \text{otherwise}.
    \end{cases}
\end{equation}

where:
\begin{itemize}
    \item \(L_{\text{SNR}}\): SNR loss, ensuring audio perceptual quality.
    \item \(L_{\text{identity}}\): Target-based loss, disrupting accurate speaker embedding extraction.
    \item \(\tau_{\text{identity}}\): Threshold determining sufficient distortion of speaker embeddings.
    \item \(\tau_{\text{SNR}}\): Threshold ensuring audio perceptual quality.
    \item \(\alpha\): Hyperparameter controlling relative emphasis between audio quality and embedding distortion.
\end{itemize}

Extensive experiments were conducted to determine optimal thresholds ($\tau_{\text{SNR}}$, $\tau_{\text{identity}}$), ensuring a balanced trade-off between adversarial robustness and perceptual audio quality. Consequently, our proposed approach generates high-quality protected audio capable of effectively resisting speech synthesis-based attacks, demonstrating practical applicability.

\begin{algorithm}[t]
\caption{Embedding-Level PGD Adversarial Perturbation}
\label{alg:pgd_attack}
\begin{algorithmic}[1]

\Require Original Voice Embedding $e_{\text{orig}}$, Target Voice Embedding $e_{\text{target}}$, Perturbation Budget $\epsilon$, Step Size $\alpha$, Number of Iterations $N$, Thresholds $\tau_{\text{identity}}$, $\tau_{\text{SNR}}$, Target Speaker Encoder $G(\cdot)$, 
\Ensure Perturbed Embedding $e_{\text{adv}}$

\State \textbf{Initialize:} $e_{\text{perturb}} \gets e_{\text{orig}} + \epsilon \cdot \mathcal{N}(0, 1)$

\For{$i = 1$ to $N$}
    \State Compute losses:
    \State \hspace{1em} $L_{\text{identity}} \gets \| G(e_{\text{perturb}})-G(e_{\text{target}})\|_{2}$

    \State \hspace{1em} $L_{\text{SNR}} \gets 10 \cdot \log_{10} \left( \dfrac{\|e_{\text{orig}}\|^2}{\|e_{\text{orig}} - e_{\text{perturb}}\|^2 + \epsilon} \right)$

    \State Compute total loss:
    \State \hspace{1em} $
        L_{\text{total}} \gets 
        \begin{cases}
            L_{\text{identity}}, & \text{if } L_{\text{identity}} \geq \tau_{\text{identity}} \\[4pt]
            L_{\text{SNR}}, & \text{if } L_{\text{identity}} < \tau_{\text{identity}} \land L_{\text{SNR}} > \tau_{\text{SNR}} \\[4pt]
            0, & \text{otherwise}
        \end{cases}
    $

    \State Compute gradient: $\nabla L_{\text{total}} \gets \dfrac{\partial L_{\text{total}}}{\partial e_{\text{perturb}}}$
    \State Update perturbation: $e_{\text{perturb}} \gets e_{\text{perturb}} + \alpha \cdot \text{sign}(\nabla L_{\text{total}})$

    \If{$L_{\text{identity}} < \tau_{\text{identity}}$ \textbf{and} $L_{\text{SNR}} \leq \tau_{\text{SNR}}$}
        \State $e_{\text{adv}} \gets e_{\text{perturb}}$
        \State \textbf{break}
    \EndIf
\EndFor

\State \Return $e_{\text{adv}}$

\end{algorithmic}
\end{algorithm}

\section{Experimental Setup}\label{sec 5}

\subsection{Speech Synthesis Models}
In this study, we employed four high-performance DNN-based speech synthesis models to effectively simulate speech synthesis attacks and evaluate the defense performance of RoVo against these attacks. These models accurately replicate the techniques used by real attackers to generate unauthorized speech through speech synthesis systems, making them essential for assessing the effectiveness of the defense mechanism. This setup enables the creation of an experimental environment that closely reflects the quality of synthetic speech and the realism of attacks.\\

\noindent \textbf{SV2TTS} \cite{jia2018transfer, jemine2019realvoice} is a deep learning model designed for voice cloning. It uses three stages: first, it creates a speaker embedding\cite{wan2018generalized} from a short audio sample; then, it generates synthetic speech from text input using this embedding. SV2TTS is effective for real-time voice synthesis and simulating voice spoofing attacks.\\

\noindent \textbf{YourTTS} \cite{casanova2022yourtts} is a zero-shot, multi-speaker TTS system introduced in 2022 that provides multilingual synthesis capabilities. This system employs an H/ASP model for speaker encoding \cite{heo2020clova}, a VITS (Variational Inference Text-to-Speech) decoder, and a HiFi-GAN vocoder \cite{kong2020hifi}. In this study, we utilized the YourTTS model from the Coqui library, based on the latest advancements in TTS research.\\

\noindent \textbf{AdaptVC} \cite{chou2019one, qian2019autovc} features an autoencoder architecture with a custom DNN utilizing instance normalization layers. This model, trained on the CSTR VCTK corpus, facilitates adaptive voice conversion across a wide range of speakers.\\

\noindent \textbf{Tortoise} TTS \cite{betker2023better, betker2022tortoise} is designed for high-quality speech synthesis, using a Transformer-based decoder to generate high-resolution audio from text. The model has been trained on a variety of speakers and speaking styles, excelling in producing natural speech from text inputs. For this study, the Tortoise TTS model was used in a \textbf{Black-box} setting to assess its performance.

\subsection{Speech Enhancment Models}
In this study, we address the issue that existing methods injecting perturbations at the signal level, which may show high defense success rates, are easily neutralized by speech enhancement techniques. These techniques can act as secondary attacks, effectively removing the injected perturbations, thereby weakening the defense performance. To evaluate this,  we simulate secondary attacks using three speech enhancement models and evaluate whether RoVo can maintain robust defense performance against these attacks.\\

\noindent \textbf{Spectral Masking} \cite{xu2014regression} operates in the frequency domain, preserving important frequency information while suppressing unnecessary frequencies. This method is effective in improving speech intelligibility and is commonly used in real-time speech enhancement applications. \\

\noindent \textbf{DeepFilterNet} \cite{schroter2022deepfilternet} is a deep learning-based adaptive filtering model that effectively separates noise from speech signals, enhancing clarity. The model performs stably in diverse environments, making it well-suited for real-time applications.\\

\noindent \textbf{MP-SENet} \cite{lu2023mp} is a multi-path attention-based network that focuses on improving speech clarity and efficiently removing noise, even in varied conditions. It has been widely recognized for its performance and versatility in handling different noise environments.

For this study, we employed both the \textbf{\_DNS} and \textbf{\_VB} versions of MP-SENet. The \textbf{\_DNS} version was trained on the Deep Noise Suppression (DNS) dataset, designed to handle a wide range of noisy conditions. The \textbf{\_VB} version was trained on the Voice Bank (VB) dataset, which contains high-quality speech recorded in controlled environments, focusing on speech restoration.

\begin{table}[t]
\centering
\caption{Example content used for synthetic speech.} \label{tab1}
\label{tab:content_examples}
\renewcommand{\arraystretch}{1.5} 
\begin{tabular}{p{0.95\linewidth}}
\hline
\textbf{Example Content} \\ \hline
``There has been a severe emergency situation, and all residents are required to evacuate immediately." \\ 
``Your credit card was used in a suspicious transaction. Can you confirm if this was authorized?" \\ 
``Your account has been temporarily locked after several failed logins. Please confirm your details to unlock it." \\ \hline
\end{tabular}
\end{table}

\subsection{Speech Corpus Dataset and Content}
In this study, we utilized the VCTK dataset \cite{veaux2017cstr}, comprising recordings from 109 English-speaking individuals. Each speaker recorded approximately 400 sentences across various content types, including news, conversational speech, and literary texts. The dataset contains diverse speech characteristics, such as intonation, pronunciation, and timbre, making it widely used in speech synthesis research. For this experiment, we used this dataset as the victim's voice. The original speech consisted of 10 samples per speaker (3–5 seconds each), totaling 1,090 samples. Adversarial perturbations were injected into the original voice to generate 1,090 defense samples, which were used in the speech synthesis models for unauthorized synthesis attacks. To evaluate the effectiveness of the defense, we generated 5,045 samples from each model based on the 1,090 defense samples. These generated samples were used to assess the defense performance against the speech synthesis models.

Additionally, to evaluate robustness against speech enhancement, the 1,090 defense samples were processed through various enhancement models. Then, 5,045 samples were generated again using the respective speech synthesis models. This process simulated a secondary attack via enhancement models to assess whether the defense maintained robustness against speech synthesis models under enhancement interference.\\

\noindent \textbf{This study} highlights the potential misuse of speech synthesis not only in terms of voice imitation but also through the generation of malicious content. Since speech synthesis models can produce TTS from text, even content the speaker never uttered, it raises real-world concerns about deepfake threats. To simulate such threats realistically, we adapted and modified content from prior studies such as Antifake. As a result, the generated speech includes content designed to trigger privacy violations or spread misinformation, with examples provided in Table \ref{tab1}.

\subsection{Evaluation Metrics}
In this study, two key metrics, \textbf{DSR (Defense Success Rate)} and \textbf{MOS (Mean Opinion Score)}, were used to evaluate the performance of the proposed defense mechanism \\

\noindent \textbf{DSR (Defense Success Rate)} is a crucial metric for assessing the effectiveness of the proposed defense method. It measures the percentage of synthesized audio, generated from adversarially perturbed speech, that is incorrectly recognized as the original speaker by a Speaker Verification model. If the synthesized audio is not recognized as the registered speaker, the defense is considered successful; otherwise, it is deemed unsuccessful. A higher DSR indicates a more effective defense mechanism, demonstrating its ability to prevent the synthesized audio from being mistaken for the original speaker. This metric critically evaluates defense performance and robustness against various synthesis models and speech enhancement techniques. \\

\noindent \textbf{MOS (Mean Opinion Score)} is a standard metric for evaluating the quality of synthesized speech, where human raters assess the naturalness and overall quality on a scale from 1 to 5. In this study, the NISQA (Neural Intrusive/Non-Intrusive Speech Quality Assessment) deep learning-based system 
 \cite{mittag2021nisqa} was used to calculate the MOS scores. A higher MOS score indicates minimal degradation of speech quality and preservation of naturalness. MOS serves as a vital indicator to assess the defense mechanism's ability to maintain high-quality audio while delivering effective defense performance. \\

\noindent  In this study, to evaluate the performance of the proposed defense mechanism, three Speaker Verification systems were utilized. The selected systems include a commercial platform and two DNN-based open-source models.

The key characteristics of each system are as follows:

\begin{itemize}
    \item \textbf{MS Azure} \cite{microsoftSpeakerRecognitionAPI} is a commercial platform for real-time speech recognition 
    and speaker verification, widely used for adversarial defense and speech synthesis evaluations.

    \item \textbf{Resemblyzer} \cite{resemblyzer2019} is a DNN-based speaker embedding model with robust verification 
    performance, frequently employed in deepfake detection and adversarial research.

    \item \textbf{ECAPA-TDNN} \cite{desplanques2020ecapa} is a high-performance DNN speaker verification model 
    generating quality embeddings, commonly applied in adversarial defense.

\end{itemize}

To ensure accurate DSR measurement for each speaker verification system, thresholds for determining whether a sample corresponds to the speaker were set using the VCTK dataset. This dataset, containing diverse speaker samples, established reliable thresholds, enabling accurate evaluation of each model’s performance.\\
\\
\\
\\
\\

\section{Experimental Results}\label{sec 6}

\textbf{\subsection{Defense Performance: RoVo}}

In this study, we evaluated the performance of RoVo, a proposed defense method designed to protect speech data against unauthorized speech synthesis attacks.We compared original unprotected (RAW) and RoVo-protected audio samples to evaluate DSR across different speech synthesis models.

The experimental results are presented in Table \ref{tab2}. Evaluations using ECAPA-TDNN showed substantial DSR improvements after applying RoVo. Specifically, the DSR increased from 22.4\% to 89.2\% for SV2TTS, from 7.16\% to 86.3\% for YourTTS, and from 26.5\% to 79.2\% for AVC. Using Resemblyzer also demonstrated clear improvements, with the DSR rising from 0.7\% to 84.9\% for SV2TTS, from 0.6\% to 70.7\% for YourTTS, and from 1.4\% to 82.8\% for AVC. Notably, tests conducted on the Microsoft Azure platform recorded exceptionally high DSR results exceeding 98\% for all speech synthesis models, highlighting RoVo’s robust defense capability.

Overall, RoVo provided an average DSR improvement of over 70\% compared to the original, unprotected audio. This demonstrates RoVo’s effectiveness in generating audio distinctly different from the original voice, effectively preventing misidentification as authentic speech. The results confirm that RoVo is a highly effective proactive measure against voice cloning attacks, significantly enhancing speech data security.

\begin{table}[t]
\caption{DSR(\%): RAW (unprotected) vs. RoVo (protected)}\label{tab2}
\centering
\begin{tabular}{lllllll}
\hline
            & \multicolumn{2}{c}{SV2TTS}                                  & \multicolumn{2}{c}{YourTTS}                                 & \multicolumn{2}{c}{AVC}                                     \\ \cline{2-7} 
            & \multicolumn{1}{c}{RAW} & \multicolumn{1}{c}{\textbf{RoVo}} & \multicolumn{1}{c}{RAW} & \multicolumn{1}{c}{\textbf{RoVo}} & \multicolumn{1}{c}{RAW} & \multicolumn{1}{c}{\textbf{RoVo}} \\ \hline
ECAPA-TDNN  & 22.4                    & \textbf{89.2}                     & 7.16                    & \textbf{86.3}                     & 26.5                    & \textbf{79.2}                     \\ \hline
Resemblyzer & 0.7                     & \textbf{84.9}                     & 0.6                     & \textbf{70.7}                     & 1.4                     & \textbf{82.8}                     \\ \hline
MS Azure    & 68.1                    & \textbf{99.8}                     & 18                      & \textbf{99.5}                     & 52.3                    & \textbf{98.6}                     \\ \hline
\end{tabular}
\end{table}

\begin{table*}[t]
\centering
\caption{DSR(\%) after applying speech enhancement models for Antifake and RoVo. Parentheses indicate the change in DSR values following enhancement.}\label{tab3}
\begin{tabular}{clllllll}
\hline
\multirow{2}{*}{Speech Enhancement} & \multirow{2}{*}{Verification} & \multicolumn{2}{c}{SV2TTS}                                             & \multicolumn{2}{c}{YourTTS}                                                     & \multicolumn{2}{c}{AVC}                                               \\ \cline{3-8} 
                                    &                               & \multicolumn{1}{c|}{Antifake}     & \multicolumn{1}{c|}{\textbf{RoVo}} & \multicolumn{1}{c|}{Antifake}     & \multicolumn{1}{c|}{\textbf{RoVo}}          & \multicolumn{1}{c|}{Antifake}     & \multicolumn{1}{c}{\textbf{RoVo}} \\ \hline
\multirow{3}{*}{Spectral Masking}   & ECAPA-TDNN                    & \multicolumn{1}{l|}{86.3 (-12.1)} & \multicolumn{1}{l|}{77.1 (-12.1)}  & \multicolumn{1}{l|}{75.6 (-17.1)} & \multicolumn{1}{l|}{85.5 (-0.8)}            & \multicolumn{1}{l|}{67.5 (-19.1)} & \textbf{83.9 (+4.7)}              \\ \cline{2-8} 
                                    & Resemblyzer                   & \multicolumn{1}{l|}{69.5 (-22.9)} & \multicolumn{1}{l|}{57.3 (-27.6)}  & \multicolumn{1}{l|}{49.3 (-32.7)} & \multicolumn{1}{l|}{66.3 (-4.4)}            & \multicolumn{1}{l|}{69.1 (-19.3)} & \textbf{90.0 (+7.2)}              \\ \cline{2-8} 
                                    & MS Azure                      & \multicolumn{1}{l|}{99.6 (-0.4)}  & \multicolumn{1}{l|}{99.0 (-0.8)}   & \multicolumn{1}{l|}{95.5 (-4.3)}  & \multicolumn{1}{l|}{98.7 (-0.8)}            & \multicolumn{1}{l|}{95.5 (-4.1)}  & \textbf{98.7 (+0.1)}              \\ \hline
\multirow{3}{*}{DeepfilterNet}      & ECAPA-TDNN                    & \multicolumn{1}{l|}{65.2 (-33.2)} & \multicolumn{1}{l|}{79.8 (-9.4)}   & \multicolumn{1}{l|}{76.9 (-15.8)} & \multicolumn{1}{l|}{84.1 (-2.2)}            & \multicolumn{1}{l|}{61.5 (-25.1)} & 79.0 (-0.2)                       \\ \cline{2-8} 
                                    & Resemblyzer                   & \multicolumn{1}{l|}{16.4 (-76.0)} & \multicolumn{1}{l|}{58.2 (-26.7)}  & \multicolumn{1}{l|}{53.0 (-29.0)} & \multicolumn{1}{l|}{\textbf{96.3 (+25.6)}}  & \multicolumn{1}{l|}{26.6 (-61.8)} & \textbf{84.4 (+1.6)}              \\ \cline{2-8} 
                                    & MS Azure                      & \multicolumn{1}{l|}{98.8 (-1.2)}  & \multicolumn{1}{l|}{99.3 (-0.5)}   & \multicolumn{1}{l|}{97.0 (-2.8)}  & \multicolumn{1}{l|}{\textbf{99.7 (+0.2)}}   & \multicolumn{1}{l|}{95.1 (-4.5)}  & 98.2 (-0.4)                       \\ \hline
\multirow{3}{*}{MP-SENet\_DNS}      & ECAPA-TDNN                    & \multicolumn{1}{l|}{49.1 (-49.3)} & \multicolumn{1}{l|}{77.8 (-11.4)}  & \multicolumn{1}{l|}{58.3 (-34.4)} & \multicolumn{1}{l|}{\textbf{89.1 (+2.8)}}   & \multicolumn{1}{l|}{53.6 (-33.0)} & 76.6 (-2.6)                       \\ \cline{2-8} 
                                    & Resemblyzer                   & \multicolumn{1}{l|}{3.0 (-89.4)}  & \multicolumn{1}{l|}{51.5 (-33.4)}  & \multicolumn{1}{l|}{25.0 (-57.0)} & \multicolumn{1}{l|}{68.7 (-2.0)}            & \multicolumn{1}{l|}{9.0 (-79.4)}  & \textbf{87.1 (+4.3)}              \\ \cline{2-8} 
                                    & MS Azure                      & \multicolumn{1}{l|}{95.8 (-4.2)}  & \multicolumn{1}{l|}{98.8 (-1.0)}   & \multicolumn{1}{l|}{89.8 (-10.0)} & \multicolumn{1}{l|}{\textbf{99.99 (+0.49)}} & \multicolumn{1}{l|}{93.9 (-5.7)}  & 98.3 (-0.3)                       \\ \hline
\multirow{3}{*}{MP-SENet\_VB}       & ECAPA-TDNN                    & \multicolumn{1}{l|}{84.0 (-14.4)} & \multicolumn{1}{l|}{81.5 (-7.7)}   & \multicolumn{1}{l|}{69.3 (-23.4)} & \multicolumn{1}{l|}{84.1 (-2.2)}            & \multicolumn{1}{l|}{66.4 (-20.2)} & \textbf{81.3 (+2.1)}              \\ \cline{2-8} 
                                    & Resemblyzer                   & \multicolumn{1}{l|}{48.7 (-43.7)} & \multicolumn{1}{l|}{64.7 (-20.2)}  & \multicolumn{1}{l|}{48.3 (-43.7)} & \multicolumn{1}{l|}{60.7 (-10.0)}           & \multicolumn{1}{l|}{20.5 (-67.9)} & \textbf{84.6 (+1.8)}              \\ \cline{2-8} 
                                    & MS Azure                      & \multicolumn{1}{l|}{99.3 (-0.7)}  & \multicolumn{1}{l|}{99.3 (-0.5)}   & \multicolumn{1}{l|}{93.0 (-6.8)}  & \multicolumn{1}{l|}{\textbf{99.6 (+0.1)}}   & \multicolumn{1}{l|}{96.5 (-3.1)}  & \textbf{98.9 (+0.3)}              \\ \hline
\end{tabular}
\end{table*}


\begin{table*}[t]
\caption{MOS comparison of Antifake and RoVo, evaluating audio quality before and after applying speech enhancement.}\label{tab4}
\centering
\begin{tabular}{cllllll}
\hline
\multirow{2}{*}{Speech enhancement} & \multicolumn{2}{c}{SV2TTS}                                       & \multicolumn{2}{c}{YourTTS}                                        & \multicolumn{2}{c}{AVC}                                            \\ \cline{2-7} 
                                    & \multicolumn{1}{c}{Antifake} & \multicolumn{1}{c}{\textbf{RoVo}} & \multicolumn{1}{c}{Antifake} & \multicolumn{1}{c}{\textbf{RoVo}}   & \multicolumn{1}{c}{Antifake} & \multicolumn{1}{c}{\textbf{RoVo}}   \\ \hline
None                                & 2.26 ± 0.51                  & 2.09 ± 0.27                       & 1.86 ± 0.40                  & {\ul \textit{\textbf{1.97 ± 0.27}}} & 2.06 ± 0.30                  & {\ul \textit{\textbf{2.91 ± 0.67}}} \\ \hline
Spectral Masking                    & 4.14 ± 0.49                  & \textbf{2.62 ± 0.84}              & 3.91 ± 0.44                  & \textbf{2.37 ± 0.77}                & 3.91 ± 0.44                  & \textbf{2.71 ± 0.83}                \\ \hline
DeepfilterNet                       & 3.98 ± 0.55                  & \textbf{3.16 ± 0.83}              & 4.28 ± 0.51                  & \textbf{3.25 ± 0.73}                & 4.24 ± 0.41                  & \textbf{3.33 ± 0.79}                \\ \hline
MP-SENet\_DNS                       & 3.87 ± 0.57                  & \textbf{3.13 ± 0.95}              & 4.02 ± 0.32                  & \textbf{3.39 ± 0.78}                & 4.11 ± 0.55                  & \textbf{3.11 ± 0.84}                \\ \hline
MP-SENet\_VB                        & 3.48 ± 0.66                  & \textbf{2.72 ± 0.85}              & 3.59 ± 0.57                  & \textbf{2.03 ± 0.72}                & 3.71 ± 0.34                  & \textbf{3.11 ± 0.74}                \\ \hline
\end{tabular}
\end{table*}

\subsection{Robustness Against Speech Enhancement}

In this study, we considered scenarios where speech enhancement models could neutralize defense-protected audio. Even defense techniques with high DSR lose their practical effectiveness if easily negated by speech enhancement models. Therefore, we experimentally verified whether RoVo provides stronger robustness against speech enhancement compared to the existing Antifake method.

As shown in Table \ref{tab3}, RoVo consistently maintained higher defense performance compared to Antifake when tested against four different speech enhancement models: Spectral Masking, DeepFilterNet, MP-SENet\_DNS, and MP-SENet\_VB. Notably, when DeepFilterNet was applied, Antifake experienced a significant DSR decrease of 76.0\% (92.4\% → 16.4\%) on the SV2TTS model, whereas RoVo showed only a moderate reduction of 26.7\% (84.9\% → 58.2\%). Moreover, in some cases (YourTTS and AVC models), applying speech enhancement models actually increased RoVo’s DSR. Upon analysis, we found that speech enhancement models did not simply remove RoVo's perturbations; instead, they severely distorted the protected audio itself. Consequently, the synthesized audio input into the verification models differed greatly from the original speaker’s voice, causing verification models to reject the audio as authentic and resulting in improved DSR.

Overall, RoVo minimized the degradation of defense performance when speech enhancement models were applied, and in certain cases even enhanced its effectiveness. These results demonstrate that embedding-level perturbations, as implemented in RoVo, offer stronger resilience against speech enhancement techniques compared to traditional signal-level perturbations, confirming RoVo as a reliable proactive defense mechanism against unauthorized speech synthesis attacks. \\

\noindent \textbf{MOS Analysis.} In this study, we evaluated whether the proposed defense method, RoVo, could maintain naturalness and audio quality even after perturbations were applied, by measuring the MOS.

As shown in Table \ref{tab4}, audio samples protected by RoVo exhibited superior audio quality compared to those protected by the existing Antifake method for both YourTTS and AVC. Notably, RoVo maintained stable MOS scores even after applying speech enhancement models such as Spectral Masking, DeepFilterNet, MP-SENet DNS, and MP-SENet VB, demonstrating strong robustness. In contrast, Antifake showed significant increases in MOS scores after enhancement models were applied, indicating that most perturbations added for defense purposes were removed, resulting in artificially improved audio quality. These results highlight that embedding-level perturbations employed by RoVo are more resistant to removal by speech enhancement models compared to the signal-level perturbations used in Antifake. Additionally, as previously observed in DSR experiments, some audio samples became severely distorted after applying speech enhancement, which explains why RoVo’s MOS scores remained relatively stable.

Overall, RoVo effectively maintained audio quality and naturalness under realistic conditions while strongly resisting the removal of perturbations by speech enhancement models. Therefore, RoVo is demonstrated to be a reliable and robust proactive defense mechanism, effectively protecting against speech synthesis attacks even under secondary attacks using speech enhancement models.

\begin{table*}[t]
\caption{DSR (\%) for Defense Performance and Transferability of Antifake and RoVo.}\label{tab5}
\centering
\begin{tabular}{ccccccc|cccccc}
\hline
\multicolumn{1}{l}{} & \multicolumn{6}{c|}{ECAPA-TDNN}                                                     & \multicolumn{6}{c}{Resemblyzer}                                                     \\ \cline{2-13} 
\multicolumn{1}{l}{} & \multicolumn{2}{c}{SV2TTS} & \multicolumn{2}{c}{YourTTS} & \multicolumn{2}{c|}{AVC} & \multicolumn{2}{c}{SV2TTS} & \multicolumn{2}{c}{YourTTS} & \multicolumn{2}{c}{AVC}  \\ \cline{2-13} 
\multicolumn{1}{l}{} & Antifake  & \textbf{RoVo}  & Antifake   & \textbf{RoVo}  & Antifake & \textbf{RoVo} & Antifake  & \textbf{RoVo}  & Antifake   & \textbf{RoVo}  & Antifake & \textbf{RoVo} \\ \hline
SV2TTS               & 98.4      & 89.2           & 74.0       & \textbf{94.5}  & 62.6     & \textbf{82.0} & 92.4      & 84.9           & 88.3       & \textbf{90.1}  & 34.3     & \textbf{66.0} \\ \hline
YourTTS              & 28.5      & \textbf{61.8}  & 92.7       & 86.3           & 28.3     & \textbf{55.8} & 3.2       & \textbf{53.2}  & 82.0       & 70.7           & 2.0      & \textbf{23.0} \\ \hline
AVC                  & 84.5      & 70.9           & 85.4       & \textbf{88.2}  & 86.6     & 79.2          & 84.5      & 66.5           & 88.3       & 66.7           & 88.4     & 82.8          \\ \hline
Tortoise             & 81.2      & \textbf{81.5}  & 84.3       & \textbf{95.3}  & 77.9     & 76.5          & 65.6      & \textbf{82.9}  & 75.2       & \textbf{90.2}  & 56.4     & \textbf{83.6} \\ \hline
\end{tabular}
\end{table*}

\subsection{Transferability: RoVo vs. Antifake}

In this study, we compared our proposed defense method, RoVo, with an existing defense method, Antifake, by evaluating their DSR and transferability. Table \ref{tab5} summarizes these evaluation results, where the second row represents the speaker encoders (speech synthesis models) used to generate defense samples, and the first column represents the black-box speech synthesis models used to evaluate those defense samples.\\

\noindent \textbf{DSR Performance Comparison.} In certain white-box scenarios, Antifake generally showed slightly higher DSR values compared to RoVo. This might be because Antifake directly injects perturbations at the audio signal level, causing a more immediate impact on the synthesized audio. However, the performance gap between Antifake and RoVo was not substantial. Thus, RoVo, which injects perturbations at the embedding level, provides a comparable level of effective defense.\\

\noindent \textbf{Transferability Comparison.}
In contrast, RoVo demonstrated notably stronger transferability than Antifake. Particularly in black-box scenarios, RoVo exhibited higher generalization and superior defense performance. For instance, defense samples generated using YourTTS achieved DSRs of 88.2\% and 95.3\% when evaluated on AVC and Tortoise, respectively, outperforming Antifake, which achieved 85.4\% and 84.3\%. In evaluations conducted with Resemblyzer, RoVo consistently showed higher DSR across multiple synthesis models. Specifically, for defense samples tested with Tortoise, RoVo achieved DSRs of 82.9\% (SV2TTS), 90.2\% (YourTTS), and 83.6\% (AVC), significantly surpassing Antifake’s 65.6\%, 75.2\%, and 56.4\%. \\


\begin{table}
\caption{DSR (\%) for Ensemble-Based Defense Performance in a Black-Box Environment.}\label{tab6}
\resizebox{0.5\textwidth}{!}{
\begin{tabular}{cllll}
\hline
                 & \multicolumn{4}{c}{\textbf{Tortoise(Black-Box)}}                                                                                                                           \\ \cline{2-5} 
                 & \multicolumn{2}{c|}{ECAPA-TDNN}                                                & \multicolumn{2}{c}{Resemblyzer}                                       \\ \cline{2-5} 
                 & \multicolumn{1}{c|}{Antifake}     & \multicolumn{1}{c|}{\textbf{RoVo}}         & \multicolumn{1}{c|}{Antifake}     & \multicolumn{1}{c}{\textbf{RoVo}} \\ \hline
Spectral Masking & \multicolumn{1}{l|}{75.1 (-18.2)} & \multicolumn{1}{l|}{\textbf{84.0 (-7.2)}}  & \multicolumn{1}{l|}{37.7 (-47.7)} & \textbf{68.8 (-3.6)}              \\ \hline
DeepfilterNet    & \multicolumn{1}{l|}{60.7 (-32.6)} & \multicolumn{1}{l|}{\textbf{80.0 (-11.2)}} & \multicolumn{1}{l|}{21.6 (-63.8)} & \textbf{48.9 (-23.5)}             \\ \hline
MP-SENet\_DNS    & \multicolumn{1}{l|}{42.9 (-40.4)} & \multicolumn{1}{l|}{\textbf{84.7 (-6.5)}}  & \multicolumn{1}{l|}{4.5 (-80.9)}  & \textbf{62.1 (-10.3)}             \\ \hline
MP-SENet\_VB     & \multicolumn{1}{l|}{55.7 (-37.6)} & \multicolumn{1}{l|}{\textbf{82.0 (-9.2)}}  & \multicolumn{1}{l|}{18.0 (-67.4)} & \textbf{41.4 (-31.0)}             \\ \hline
\end{tabular}
}
\end{table}

\noindent \textbf{Ensemble Defense in Black-Box Settings.} Antifake is known to achieve high DSR in black-box scenarios through ensemble training that utilizes multiple speaker encoders. A notable advantage of ensemble-based methods is their capability to generalize effectively, providing robust protection against previously unseen models. To evaluate this ensemble effect, we optimized adversarial perturbations using speaker encoders from three speech synthesis models (SV2TTS, YourTTS, AVC). Subsequently, we evaluated the defense performance on the Tortoise model, which was entirely excluded from the training process, simulating a black-box.

To ensure fair performance comparisons, RoVo was optimized without the PerC-AL framework, employing the same single-loss optimization method as Antifake. Our experiments particularly focused on assessing robustness under secondary attacks involving speech enhancement techniques. The experimental results are summarized in Table \ref{tab6}. Initial defense success rates (before speech enhancement) showed that Antifake slightly outperformed or was comparable to RoVo in absolute terms. However, the performance drop (shown in parentheses in Table \ref{tab6}) after applying speech enhancement techniques was significantly smaller for RoVo compared to Antifake. This clearly indicates that embedding-level perturbations are fundamentally more challenging for speech enhancement models to neutralize. For instance, using the ECAPA-TDNN with MP-SENet\_DNS enhancement, Antifake experienced a substantial performance drop of 40.4\%, whereas RoVo showed only a minimal drop of 6.5\%, thus maintaining stronger defense robustness. These results demonstrate that RoVo’s embedding-level perturbation approach provides superior generalization through ensemble training and exhibits higher resilience to post-processing attacks compared to conventional signal-level approaches such as Antifake.

Overall, while RoVo achieved slightly lower or comparable DSR values compared to Antifake in certain white-box scenarios, it consistently demonstrated superior robustness and generalization performance in ensemble-based black-box evaluations and transferability assessments. Notably, RoVo maintained significantly higher robustness against secondary attacks, such as speech enhancement, due to its embedding-level perturbation strategy, clearly outperforming the signal-level perturbation approach employed by Antifake. These results highlight that RoVo provides strong generalization and reliable proactive defense across diverse unseen synthesis models and realistic threat environments, thereby establishing RoVo as a highly effective and practical voice protection mechanism in real-world applications.

\subsection{User Study}

In this study, we conducted a user study to evaluate whether our proposed defense mechanism, RoVo, can effectively protect voices from unauthorized speech synthesis attacks while maintaining naturalness and audio quality. Our experimental design was based on widely-used methodologies from related studies to ensure consistency and comparability. The evaluation involved 100 native English speakers of various ages recruited through the Prolific platform\cite{prolific2025}. Participants spent approximately 10 minutes on the task and received a compensation of \$2.
Participants listened to a total of 30 pairs of audio samples and rated how similar each pair sounded, indicating whether they believed the samples originated from the same speaker. Responses were recorded using a 5-point Likert scale: "Very Similar," "Somewhat Similar," "Neutral," "Somewhat Different," and "Very Different." The audio pairs were constructed using six distinct combinations:

\begin{itemize}
    \item \textbf{RA\_RA (Real A/Real A)}: Two natural utterances from the same speaker.

    \item \textbf{RA\_RB (Real A/Real B)}: Two natural utterances from different speakers.
    \item \textbf{RA\_DA (Real A/Defense A)}: A defended utterance from the same speaker.
    \item \textbf{RA\_FDA (Real A/Fake A)}: A defended synthesized utterance.
    \item \textbf{RA\_DA-SE (Defense A with Speech Enhancment)}
    \item \textbf{RA\_FDA-SE (Fake A with Speech Enhancment)}
    
\end{itemize}

\begin{figure}[t]
\centering
\includegraphics[width=0.45\textwidth]{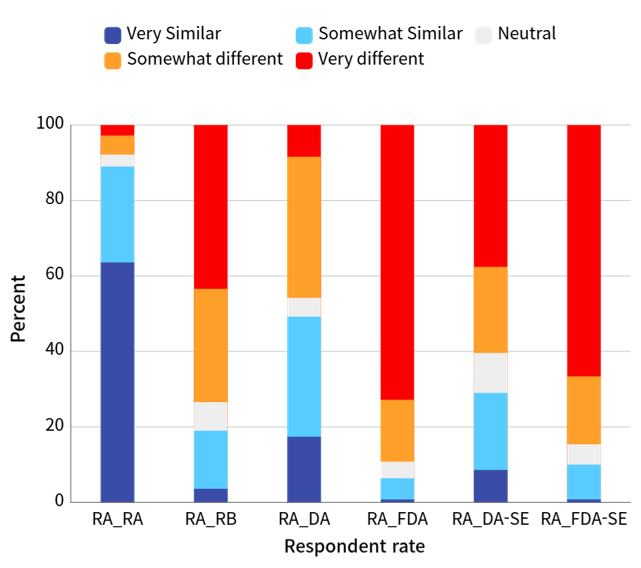}
\caption{User Study Results: Comparison of Perceived Similarity Before and After Speech Enhancement.} 
\label{fig_5}
\end{figure}

\noindent The experimental results are presented in Figure \ref{fig_5}.\\
\noindent \textbf{RA\_RA \& RA\_RB.}
First, comparisons between original reference audios (RA\_RA) received predominantly "Very Similar" ratings from participants, confirming the reliability of our evaluation methodology. Conversely, comparisons involving clearly distinct speakers (RA\_RB) were mostly rated as "Very Different" or "Somewhat Different," further validating the test criteria.\\

\noindent \textbf{RA\_DA \& RA\_DA\-SE.}
For RoVo-defended natural speech samples (RA\_DA), approximately 50\% of participants rated the audio as "Very Similar" or "Somewhat Similar" to the original, indicating that RoVo effectively preserves the naturalness and audio quality while providing robust defense. However, after applying speech enhancement (RA\_DA\-SE), about 80\% of respondents shifted their ratings to "Somewhat Different" or "Very Different." This indicates that speech enhancement methods, intended to remove perturbations, inadvertently introduced distortions, significantly reducing the perceived naturalness.\\

\noindent \textbf{RA\_FDA \& RA\_FDA\-SE.}
Regarding RoVo-defended synthesized speech samples (RA\_FDA), around 73\% of participants rated these as "Very Different" from the original natural audio. This clearly demonstrates RoVo's effectiveness in distinguishing synthesized from genuine speech. Even after speech enhancement was applied (RA\_FDA-SE), about 67\% still rated the audio as "Very Different," confirming that RoVo’s perturbations partially persisted, maintaining robust defensive performance against synthesis attacks.\\

\noindent \textbf{Overall Analysis.}
Overall, the user study results clearly demonstrate RoVo’s capability to robustly protect voice data from unauthorized synthesis attacks while effectively preserving the audio quality and naturalness of original speech. Furthermore, RoVo showed high resilience even under secondary speech enhancement attacks, indicating its practicality and reliability as a proactive voice-defense mechanism in real-world applications.

\section{Discussion} \label{sec 7}
In this study, we introduced RoVo, a novel proactive defense framework that injects adversarial perturbations at the embedding level to address critical limitations of existing defenses against speech synthesis attacks. RoVo provides a practical and effective solution in response to the rising malicious use of AI-based speech synthesis technologies. In this section, we discuss the strengths of our proposed approach based on extensive experimental results, as well as outline limitations and suggest future research directions.\\

\noindent \textbf{Embedding-Level Perturbations.} Our embedding-level perturbation method successfully overcomes fundamental weaknesses inherent in traditional signal-level perturbation methods. Specifically, RoVo demonstrated significantly higher robustness compared to the signal-level baseline (Antifake) when evaluated against multiple speech enhancement models. This advantage primarily arises from the high-dimensional and nonlinear nature of the embedding space, making perturbation removal challenging. However, despite this notable strength, embedding-level perturbations still inherently compromise audio perceptual quality to some extent, as shown in our MOS evaluations. Perturbation injection inevitably introduces subtle distortions compared to original speech, a fundamental limitation shared by all existing proactive defense mechanisms. Minimizing quality degradation caused by perturbations thus remains an essential avenue for future research. \\

\noindent \textbf{Performance in White-box and Black-box.} RoVo consistently exhibited robust defense performance in both white-box and black-box experimental setups. By injecting perturbations directly into the internal embedding space, RoVo effectively generalized across various attacker models. However, in certain white-box scenarios, the baseline Antifake occasionally demonstrated slightly superior defense success rates prior to speech enhancement. This suggests that signal-level perturbations can provide immediate and intuitive effectiveness, indicating potential for complementary use alongside embedding-level approaches in future research.\\ 

\noindent \textbf{Robustness to Speech Enhancement Attacks.} A major contribution of this research is experimentally demonstrating RoVo's strong robustness against secondary attacks such as speech enhancement. Unlike Antifake, whose perturbations were effectively neutralized by enhancement models, RoVo maintained substantial robustness. Moreover, speech enhancement attempts sometimes severely distorted critical features of RoVo-protected speech, reducing its usability for unauthorized synthesis—effectively serving as an additional protective mechanism.\\

\noindent \textbf{User Study: Implications and Limitations.} Our user study results confirm RoVo’s ability to maintain acceptable audio quality and naturalness while effectively safeguarding voices against unauthorized synthesis. Most participants reported high similarity between RoVo-protected audio and original speech, underscoring RoVo’s practical utility. Nonetheless, our evaluation did not comprehensively quantify potential discomfort or cognitive impact from varying perturbation intensity levels. Future studies should thus include more detailed quantitative analysis and expanded user evaluations to precisely assess RoVo’s practical usability.\\

\section{Conclusion} \label{sec 8}
In summary, our proposed RoVo framework substantially advances the state-of-the-art in proactive defense against speech synthesis attacks by demonstrating robust protection and strong practical applicability. Although further research is required to validate RoVo’s generalizability and resilience in diverse real-world scenarios, particularly considering factors such as varying audio characteristics, background noise, and efficient real-time deployment, our results provide a solid foundation toward addressing these challenges. Future work should also focus on optimizing perturbation strategies to balance defense robustness and audio quality. Building upon the limitations and insights discussed in this study, we aim to expand RoVo's effectiveness across a broader range of realistic scenarios, further enhancing its practicality and reliability. We envision RoVo as a reliable and practical proactive solution against emerging threats posed by malicious speech synthesis.

\bibliographystyle{ACM-Reference-Format}
\bibliography{RoVo}

\end{document}